\documentclass[10pt,letterpaper]{article}
\usepackage{cvpr}
\usepackage{times}
\usepackage{epsfig}
\usepackage{graphicx}
\usepackage{amsmath}
\usepackage{amssymb}
\usepackage{booktabs}
\usepackage{algorithm}
\usepackage{algorithmic}
\usepackage{float}
\usepackage{hyperref}
\usepackage{multirow}
\usepackage{subfigure}
\usepackage{CJKutf8}
\usepackage{listings}
\usepackage{xcolor}
\usepackage{enumitem}

\lstdefinestyle{promptstyle}{
  backgroundcolor=\color{gray!10},
  basicstyle=\ttfamily\footnotesize,
  frame=single,
  breaklines=true,
  framerule=0.5pt,
  rulecolor=\color{gray},
  captionpos=t
}

\graphicspath{{./}}

\title{\LARGE ODIA: Oriented Distillation for Inline Acceleration of LLM-based Function Calling}

\author{Hanlong Zhang, Jingsheng Yang, Hao Li, Yuhao He, Franck Gong\\
ByteDance Inc.\\
{\tt\small \{zhanghanlong.go, yangjingsheng, lihao.1021, yuhao.he, franck.gong\}@bytedance.com}
}

\begin{document}

\twocolumn[{
\maketitle
\vspace{1em} 
}]

\begin{abstract}
Function Calling is a crucial technique that enables Large Language Models (LLMs) to interact with external systems through APIs. However, the high latency associated with LLM-based Function Calling significantly impacts user experience. This paper presents a novel approach called Oriented Distillation for Inline Acceleration (ODIA) that leverages online user interaction data to accelerate Function Calling. By automatically identifying "simple queries" from production traffic and distilling knowledge from larger models to smaller ones, our method reduces response latency by 45\% (expected) and 78\% (median) while maintaining accuracy. We demonstrate the effectiveness of our approach through real-world deployment in a music application, where the smaller model successfully handles 60\% of traffic with negligible accuracy loss. Our method requires minimal human intervention and continuously improves through automated data collection and model updating, making it a practical solution for production environments.
\end{abstract}

\section{Introduction}
Large Language Models (LLMs) have demonstrated remarkable capabilities across various domains. Function Calling, a technique that allows LLMs to select appropriate functions and generate the necessary parameters based on user queries, has become an essential component in many applications. This capability enables LLMs to connect with external systems, such as APIs, effectively bridging the gap between natural language understanding and actionable operations.

Despite its utility, Function Calling faces a significant challenge: latency. When processing user queries, LLMs typically require 1-2 seconds to complete a Function Calling operation, with peak (P90) latencies reaching 2-3 seconds during high-traffic periods. Unlike streaming text generation, Function Calling operates as a "black box" intermediate process from the user's perspective, making wait times feel particularly long and negatively impacting user experience.

\begin{figure}[t]
\centering
\includegraphics[width=\columnwidth]{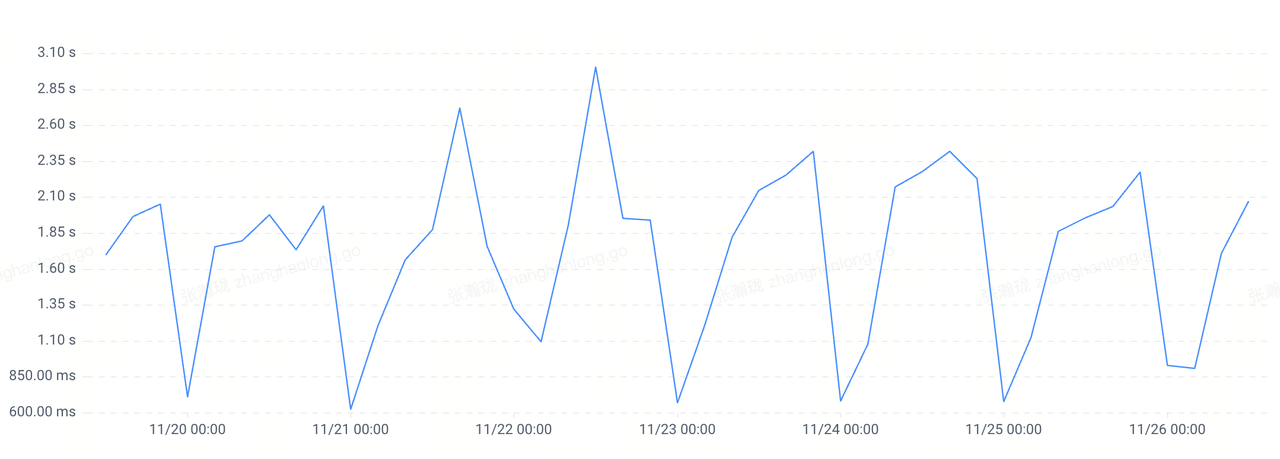}
\caption{P90 Function Calling latency reached 2–3 seconds during peak hours in the music application.}
\label{fig:fc_latency}
\end{figure}

Current approaches to accelerate LLM inference, such as quantization and optimized serving strategies, provide limited improvements for Function Calling specifically. Manual efforts to build specialized models for simple queries require significant human resources for data annotation, model training, and maintenance, making them difficult to scale across multiple domains and use cases.

In this paper, we present Oriented Distillation for Inline Acceleration (ODIA), a novel approach that leverages online user interaction data to automatically accelerate Function Calling. Our key contributions include:

\begin{itemize}
    \item A method to automatically identify "simple queries" from production traffic that smaller models can reliably handle
    \item A knowledge distillation framework that transfers Function Calling capabilities from larger to smaller models
    \item A dual-model architecture that efficiently routes queries between small and large models based on complexity
    \item Empirical validation through deployment in a production music application, demonstrating significant latency improvements while maintaining accuracy
\end{itemize}

\section{Related Work}

\subsection{Speculative Decoding}
Speculative decoding is a technique designed to accelerate text generation in large language models. The core idea involves using a smaller, faster "draft model" to predict multiple tokens, which are then verified by a larger, more accurate target model. This approach significantly reduces the computational load on the larger model, as it only needs to confirm or reject the predictions rather than generate tokens from scratch \cite{speculative, spector2023accelerating, zhang2024learning}.

The effectiveness of speculative decoding has been demonstrated in various applications, particularly in scenarios with abundant "boilerplate" text, such as code generation. However, this approach typically requires coordination between the small and large models, making it challenging to implement from a service integration perspective unless both models are managed by the same provider \cite{zhang2025swiftspec}.

Recent advances in speculative decoding have further improved its efficiency. Zhang et al. \cite{zhang2025swiftspec} introduced SwiftSpec, which redesigns the speculative decoding pipeline in an asynchronous and disaggregated manner, allowing each component to be scaled flexibly. Chen et al. \cite{chen2025spin} proposed SPIN, which improves token speculation by using multiple heterogeneous small speculative models with a learning-based algorithm for model selection. These approaches have demonstrated significant speedups in LLM inference, with performance increases of approximately 2.28x compared to standard methods \cite{chen2025spin}.

\subsection{Semantic Cache}
Semantic Cache is an intelligent caching mechanism based on semantic similarity, designed to provide faster responses for natural language tasks. Unlike traditional caching systems that rely on exact matches, Semantic Cache captures and stores the meaning of text, enabling it to retrieve relevant information even when inputs are phrased differently \cite{semantic, regmi2024gpt}.

This approach is primarily offered by vector database providers such as Zilliz and MongoDB. The fundamental concept involves storing user query-response pairs and retrieving stored responses for new queries that exceed a similarity threshold when compared to previously seen queries \cite{iyengar2025generative}.

While Semantic Cache is not limited to specific task types, it is particularly effective for answering popular single-turn questions. However, it suffers from low hit rates for extended conversations and may not meet accuracy requirements for Function Calling operations that involve write actions due to the approximate nature of semantic matching \cite{zhu2024efficient}.

Recent research has focused on improving the efficiency and effectiveness of semantic caching for LLMs. Regmi and Pun \cite{regmi2024gpt} introduced GPT Semantic Cache, which leverages semantic embedding caching in in-memory storage to identify semantically similar questions, achieving up to 68.8\% reduction in API calls across various query categories. Iyengar et al. \cite{iyengar2025generative} proposed a generative caching system that can synthesize multiple cached responses to provide answers to previously unseen queries, significantly reducing both latency and monetary costs for accessing LLMs.

\subsection{Function Calling Optimization}
Function Calling in LLMs has become increasingly important for integrating language models with external systems. Recent research has focused on improving the efficiency and reliability of this capability. Gim et al. \cite{gim2024asynchronous} proposed AsyncLM, a system for asynchronous LLM function calling that enables LLMs to generate and execute function calls concurrently, reducing end-to-end task completion latency by 1.6x-5.4x compared to synchronous function calling.

Liu et al. \cite{liu2025hamburger} introduced HAMburger, a hierarchically auto-regressive model that redefines resource allocation in LLMs by moving beyond uniform computation and storage per token during inference. This approach allows for smashing multiple tokens into a single KV cache and generating several tokens per step, reducing KV cache computation by up to 2x and achieving up to 2x tokens per second while maintaining quality.

These advancements in Function Calling optimization complement our approach by addressing different aspects of the latency challenge. While these methods focus on improving the efficiency of the function calling process itself, our work concentrates on identifying and routing "simple queries" to smaller, specialized models, thereby reducing the overall computational load and response time.

\section{Methodology}

\subsection{Overall Design}
Our approach to accelerating LLM Function Calling consists of two main components: an offline pipeline and an online serving system. The offline pipeline automatically identifies "simple queries" from production data and trains specialized models, while the online system efficiently routes incoming queries to either the small or large model based on their complexity.

\begin{figure}[t]
\centering
\includegraphics[width=0.5\textwidth]{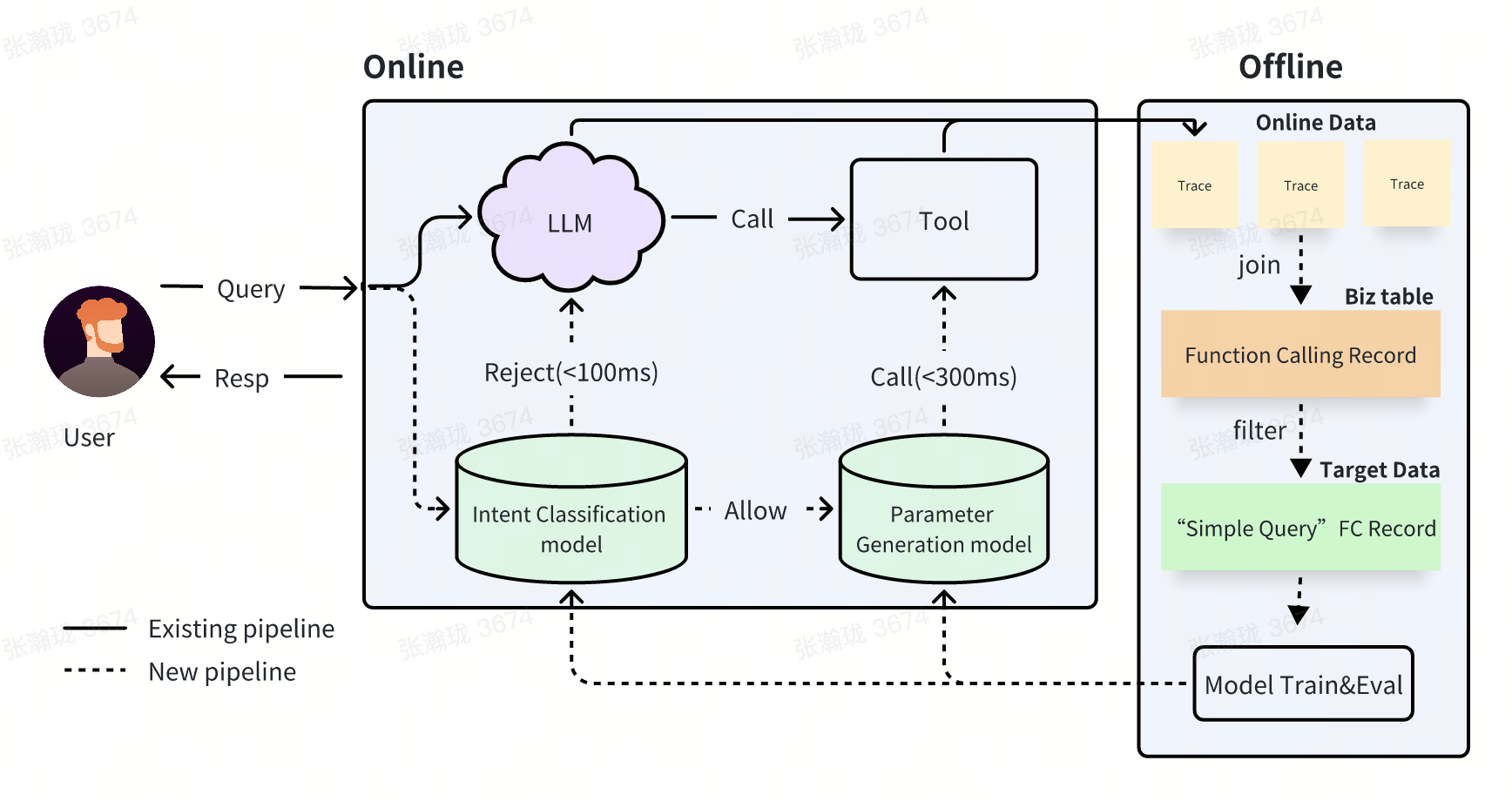}
\caption{Overall architecture of our approach, showing both offline pipeline for model training and online serving system for query routing}
\label{fig:overall_design}
\end{figure}

The overall process follows these steps:

\begin{enumerate}
    \item \textbf{Data Accumulation:} We collect Function Calling data from production traffic, including inputs (user queries, conversation history, tool definitions) and outputs (function names and parameters).
    
    \item \textbf{Automatic Filtering:} Using minimal human intervention, we algorithmically identify "simple" queries that smaller models can reliably handle.
    
    \item \textbf{Small Model Training:} We train two models using the filtered data: (1) an intent classification model that quickly determines whether a query can be handled by the small model, and (2) a parameter generation model that produces Function Calling outputs for simple queries.
    
    \item \textbf{Online Acceleration:} In production, we use the classification model to route queries, with complex queries falling back to the large model.
\end{enumerate}

\subsection{Defining "Simple" Queries}
A key challenge in our approach is defining what constitutes a "simple" query that smaller models can reliably handle. Initially, we hypothesized that queries with consistent function selection would be simpler. However, analysis revealed that even semantically similar queries might call different functions due to LLM instability and semantic complexity.

We refined our definition to focus on query clusters that demonstrate consistent function selection behavior. Specifically, we consider a cluster of semantically similar queries to be "simple" if a high percentage of them result in calls to the same function. This consistency indicates that the function selection pattern is stable and learnable by a smaller model.

For example, queries like "play Jay Chou's songs" and "play some music by Jay Chou" consistently call the same playback function, making them "simple queries." Conversely, ambiguous queries like "switch" or context-dependent queries like "more of these" may call different functions based on conversation context, marking them as "complex queries."

The following examples from our production data illustrate the distinction between simple and complex queries:

\begin{itemize}[leftmargin=1.5em]
  \item \textbf{Simple queries} (consistent function selection):

  \begin{itemize}[label={}, leftmargin=0.1em]
    \item \texttt{"Play Jay Chou's songs"} $\rightarrow$ \textit{audioSearch}
    \item \texttt{"Play some music by Jay Chou"} $\rightarrow$ \textit{audioSearch}
    \item \texttt{"I want to listen to some jazz"} $\rightarrow$ \textit{intentionRecommend}
    \item \texttt{"Give me some jazz music"} $\rightarrow$ \textit{intentionRecommend}
  \end{itemize}

  \item \textbf{Complex queries} (ambiguous or context-dependent):

  \begin{itemize}[label={}, leftmargin=0.1em]
    \item \texttt{"Switch/change"} $\rightarrow$ \textit{playControl} or \textit{intentRecommend} (depending on context)
    \item \texttt{"More of these"} $\rightarrow$ \textit{onlineSearch} or \textit{intentRecommend} (depending on previous conversation)
    \item \texttt{"Don't want to listen"} $\rightarrow$ triggers different control functions based on current state
  \end{itemize}
\end{itemize}

\subsection{Identifying Similar Queries}
To identify clusters of similar queries, we employ two complementary approaches:

\subsubsection{Semantic Similarity-Based Clustering}
We convert user queries into vector representations using embedding models optimized for Chinese paraphrase detection. After evaluating several candidates, we selected the shibing624/text2vec-base-chinese-paraphrase model for its performance in our domain.

We then apply hierarchical clustering with a similarity threshold to group queries without predetermining the number of clusters. This approach effectively groups queries with similar semantic meanings despite differences in phrasing.

Our clustering results revealed clear patterns in user queries. For example, in the music domain, we identified distinct clusters for recommendation requests, search requests, and playback control:

\begin{itemize}[leftmargin=1.5em]
  \item \textbf{Recommendation cluster} (mapped to \textit{intentionRecommend} function):

  \begin{itemize}[label={}, leftmargin=0.1em]
    \item \texttt{"I want to listen to some jazz"} $\rightarrow$ \textit{intentionRecommend}
    \item \texttt{"I'm into jazz lately, recommend more to me"} $\rightarrow$ \textit{intentionRecommend}
    \item \texttt{"Recommend jazz songs"} $\rightarrow$ \textit{intentionRecommend}
    \item \texttt{"Give me some jazz"} $\rightarrow$ \textit{intentionRecommend}
  \end{itemize}

  \item \textbf{Search cluster} (mapped to \textit{audioSearch} function):

  \begin{itemize}[label={}, leftmargin=0.1em]
    \item \texttt{"Help me play a song by Fenghuang Legend"} $\rightarrow$ \textit{audioSearch}
    \item \texttt{"Play me a song by Fenghuang Legend"} $\rightarrow$ \textit{audioSearch}
    \item \texttt{"Play a song by Fenghuang Legend"} $\rightarrow$ \textit{audioSearch}
  \end{itemize}

  \item \textbf{Control cluster} (mapped to \textit{playerControl} function):

  \begin{itemize}[label={}, leftmargin=0.1em]
    \item \texttt{"I don't want to listen"} $\rightarrow$ \textit{playerControl}
    \item \texttt{"Don't listen"} $\rightarrow$ \textit{playerControl}
    \item \texttt{"Stop, I don't want to listen"} $\rightarrow$ \textit{playerControl}
  \end{itemize}
\end{itemize}

\subsubsection{Named Entity Recognition (NER) Based Clustering}
While semantic clustering groups similar expressions, it lacks abstraction of expression patterns. For instance, \texttt{"play Fragrance of Rice"} and \texttt{"play Listen to Mom"} would be placed in different clusters despite following the same pattern.

To address this limitation, we extract entities from queries using domain-specific entity dictionaries (songs, artists, movies) and convert queries into templates. For example, \texttt{"play Jay Chou's Fragrance of Rice"} becomes \texttt{"play $<$artist$>$'s $<$song$>$"}. This allows the model to learn general patterns rather than specific instances.

We constructed comprehensive entity dictionaries for our music domain, including:
\begin{itemize}
    \item \textbf{Song dictionary}: Contains thousands of song titles
    \item \textbf{Artist dictionary}: Contains names of singers, bands, and composers
    \item \textbf{Genre dictionary}: Contains music genres and categories
\end{itemize}

This approach enabled us to identify common query patterns such as:
\begin{itemize}[label={}, leftmargin=2em]
  \item \texttt{"<artist>'s new song"}
  \item \texttt{"What songs does <artist> have"}
  \item \texttt{"<movie/TV show> theme song"}
\end{itemize}

\subsection{Filtering Simple Queries}
We organize our data hierarchically: individual Function Calling records are grouped by identical queries, which are then clustered based on semantic similarity and NER patterns.

\begin{figure}[t]
\centering
\includegraphics[width=\columnwidth]{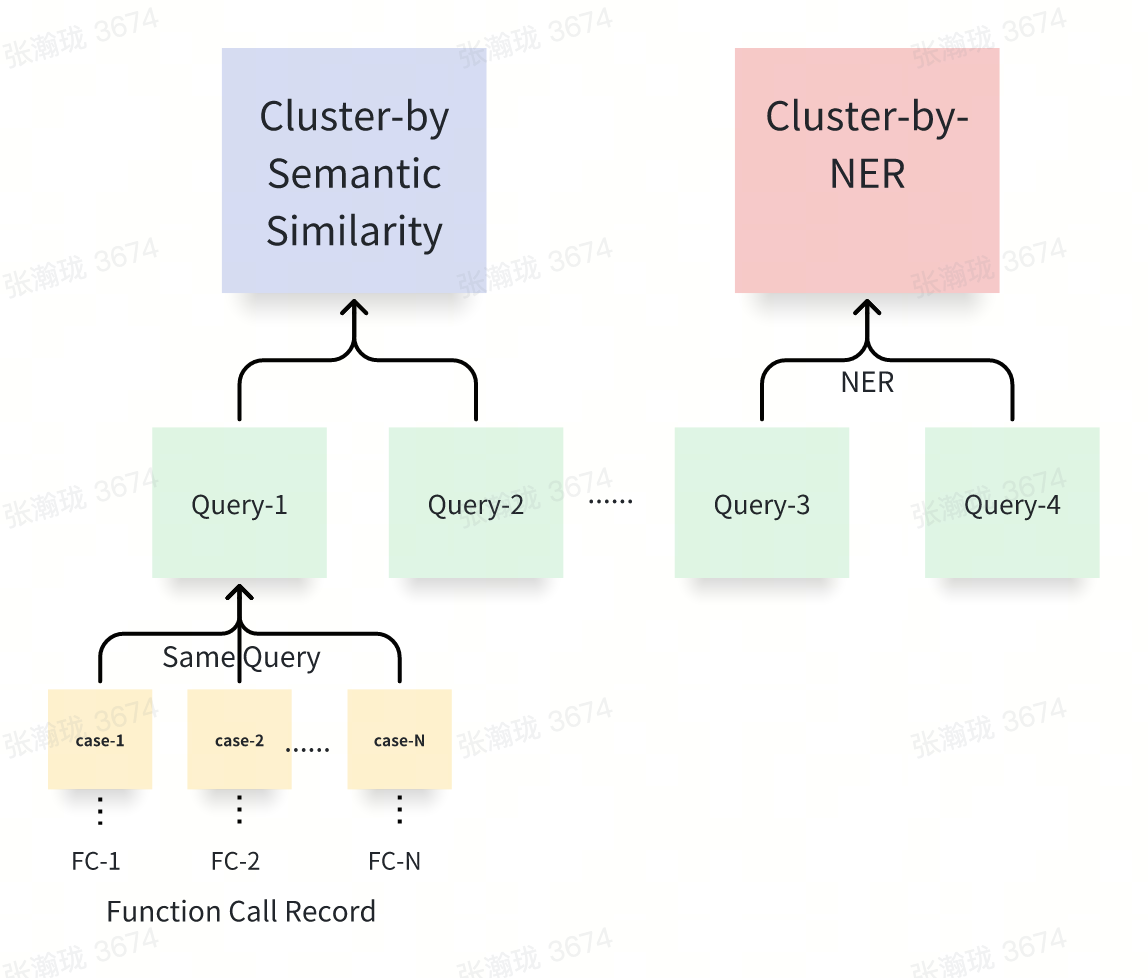}
\caption{Hierarchical data organization for filtering simple queries}
\label{fig:data_organization}
\end{figure}

To identify simple query clusters:
\begin{itemize}
    \item We calculate the frequency distribution of functions called within each cluster
    \item Clusters where the most frequent function exceeds a threshold percentage are labeled as "simple"
    \item Within these clusters, we discard individual queries that call functions other than the dominant one
    \item Clusters without a dominant function are classified as containing "complex queries"
\end{itemize}

For training data construction, we employ balanced sampling based on function distribution, downsampling over-represented functions and supplementing under represented ones. When multiple records exist for identical queries, we prioritize recent examples with more complete parameters.

\subsection{Model Architecture}
Our system employs two specialized models:

\subsubsection{Intent Routing Model}
This model determines whether an incoming query can be handled by the small model or should be routed to the large model. Given its position at the entry point of all requests, it must achieve high accuracy ($>$95\%) and low latency ($<$50ms).

We explored three approaches:
\begin{itemize}
    \item \textbf{Semantic Similarity + NER Matching:} Compare incoming queries against known clusters and templates
    \item \textbf{Traditional Classification with Embeddings:} Train models like XGBoost or Random Forest on query embeddings
    \item \textbf{BERT Classification:} Directly train a BERT model on clustered queries
\end{itemize}

Initially, we implemented the first approach for high precision, handling only previously seen queries. To improve coverage, we later added the second approach, increasing 12\% in offline evaluations.

\subsubsection{Parameter Generation Model}
This model selects appropriate tools and extracts parameters. Balancing context understanding with latency constraints ($<$300ms), we evaluated several models under 2B parameters, including deepseek-coder-1.3B, miniCPM-2B, and qinwen2-1.5B. After thorough offline evaluation of these models, we ultimately selected deepseek-coder-1.3B for its optimal balance of performance, accuracy, and inference speed.

The model input includes:
\begin{itemize}
    \item A system prompt defining the task and available functions
    \item Up to three recent conversation turns
    \item The current user query
\end{itemize}

Here's an example of the system prompt format we use:
\lstset{breaklines=true, basicstyle=\ttfamily\small}

\begin{lstlisting}[style=promptstyle]
# Role
You are a parameter extraction bot. Your task 
is to analyze the user's conversation history 
and current query, select the appropriate tool, 
and output the corresponding parameters.

# Tool Definition
{{Tool Schema in JSON format}}

# Output in the following JSON format
{"name":"tool_name","arguments":{"key1": 
"value1"}}

Based on the following conversation:
### Instruction:
Pause playback
### Response:
\end{lstlisting}

The output is a JSON-formatted Function Calling result containing the function name and arguments, for example:
\begin{lstlisting}[style=promptstyle]
{"name": "control", "arguments": {"command": 
"Pause"}}
\end{lstlisting}

\subsection{Incremental Updates}
To handle large data volumes and continuously improve coverage, we implemented an incremental update mechanism inspired by recurrent neural networks. Rather than processing all data at once, we process data in fixed-size batches (80,000-100,000 examples), allowing us to unify one-time bulk training with daily updates.

For merging new and existing clusters:
\begin{enumerate}
    \item We represent each cluster by the weighted average of its query embeddings
    \item We perform second-level clustering on these cluster representations
    \item For successfully merged clusters, we combine their queries and associated examples
    \item We apply pruning to maintain manageable cluster sizes, merging very similar queries and limiting the number of examples per query
\end{enumerate}

\begin{figure}[t]
\centering
\includegraphics[width=\columnwidth]{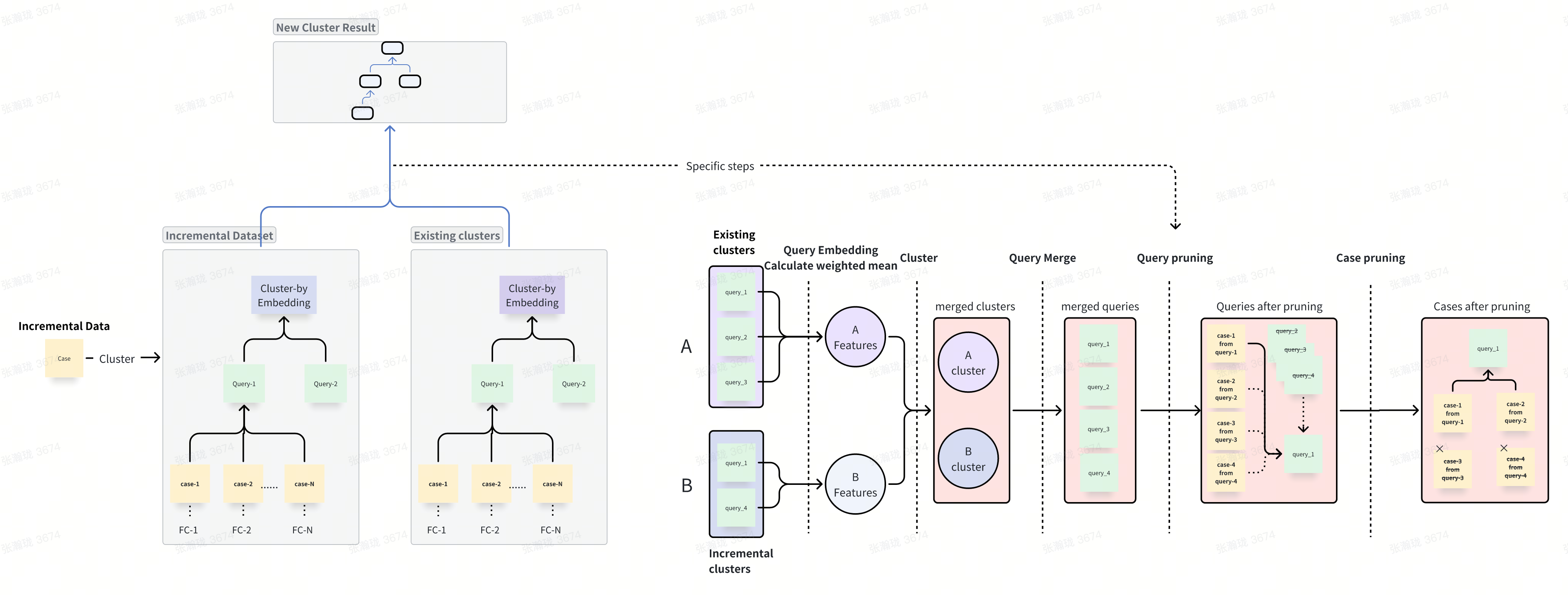}
\caption{Process of merging new and existing clusters}
\label{fig:cluster_merging}
\end{figure}

\subsection{Performance Optimization}
Beyond model substitution, we explored several techniques to further optimize small model performance:

\begin{itemize}
    \item \textbf{System Prompt Optimization:} Streamlining system prompts to reduce input token count while balancing performance
    \item \textbf{Output Simplification:} Removing common prefixes from Function Calling outputs to reduce output token length
    \item \textbf{Token Optimization:} Replacing multi-token parameter names with single-token alternatives to reduce sequence length
\end{itemize}

For example, in our parameter optimization efforts, we identified that certain parameter names in our function definitions were consuming multiple tokens, such as:

\begin{lstlisting}[style=promptstyle]
"parameters": {
  "media_type": "string",
  "creator_name": "string",
  "media_name": "string"
}
\end{lstlisting}

We optimized these by replacing multi-token parameter names with single-token alternatives:

\begin{lstlisting}[style=promptstyle]
"parameters": {
  "type": "string",
  "creator": "string",
  "media": "string"
}
\end{lstlisting}

This optimization reduced the total token count and improved inference speed without affecting the model's understanding of the parameters' meanings.

\section{Experiments}

\subsection{Experimental Results}
Our deployment demonstrated significant performance improvements:

\subsubsection{Latency Reduction}
Our experiments show that the small model processes Function Calling requests substantially faster than the large model. Visual comparison shows that while the large model is still generating its first line of output, the small model has already completed four lines.

Quantitatively, with the small model handling approximately 60\% of traffic, we observed:
\begin{itemize}
    \item Expected latency reduction of 45\% (from 1600ms to 870ms)
    \item Median latency reduction of 78\% (from 1600ms to 350ms)
\end{itemize}

The expected latency calculation follows:
\begin{equation}
E(T_{pre\_fc}) = 50 + 300 \times 60\% + 1600 \times 40\% = 870 \text{ ms}
\end{equation}
where 50ms represents routing overhead, 300ms is small model latency, and 1600ms is large model latency.

\begin{table}[h!]
\centering
\begin{tabular}{lccccc}
\toprule
GPU & QPS & P99(ms) & AVG(ms) & MAX(ms) \\
\midrule
A30 & 4   & 247.77  & 245.63  & 308     \\
L20 & 5   & 246.96  & 243.67  & 312     \\
\bottomrule
\end{tabular}
\caption{Model Performance Evaluation}
\end{table}

\subsubsection{Coverage and Accuracy}
Our intent routing model successfully identified approximately 60\% of queries as candidates for small model processing. Within this subset, the parameter generation model achieved accuracy comparable to the large model, with negligible degradation in function selection and parameter extraction.

\subsubsection{Resource Efficiency}
Beyond latency improvements, our approach significantly reduced computational resource requirements. The small model (1.3B parameters) requires substantially less GPU memory and computational power than the large model, leading to cost savings in production deployment.

\subsection{Future Plans}
Based on the success of our initial deployment, we plan several enhancements:

\begin{itemize}
    \item \textbf{Coverage Expansion:} Improving the intent routing model to safely handle a larger percentage of queries
    \item \textbf{Cross-Domain Application:} Extending the approach to other application domains beyond music
    \item \textbf{Automated Feedback Loop:} Implementing continuous learning from production data to improve both models over time
    \item \textbf{Model Architecture Optimization:} Exploring specialized architectures for Function Calling to further reduce latency
\end{itemize}

\section{Conclusion}
This paper presents a practical approach to accelerating Large Language Model Function Calling using online data. By automatically identifying "simple queries" from production traffic and distilling knowledge from larger models to smaller ones, we achieved significant latency reductions while maintaining accuracy. Our method requires minimal human intervention and continuously improves through automated data collection and model updating.

The success of our deployment in a music application demonstrates the viability of this approach for production environments. As LLMs continue to be integrated into more applications, techniques like ours that balance performance, efficiency, and accuracy will become increasingly important.

Looking forward, we believe that leveraging the vast amount of online interaction data represents a promising direction for improving LLM performance across various tasks. The bitter lesson from AI research history reminds us that approaches that can automatically exploit large amounts of data often outperform hand-engineered solutions in the long run.

\end{document}